\documentclass[letterpaper]{article} 
\usepackage{aaai25}  
\usepackage{times}  
\usepackage{helvet}  
\usepackage{courier}  
\usepackage[hyphens]{url}  
\usepackage{graphicx} 
\usepackage{natbib}  
\usepackage{caption} 
\usepackage{algorithm}
\usepackage{algorithmic}
\usepackage{url}            
\usepackage{booktabs}       
\usepackage{amsfonts}       
\usepackage{nicefrac}       
\usepackage{microtype}      
\usepackage{xcolor}         

\usepackage{graphicx}
\usepackage{soul}
\usepackage{enumitem}
\usepackage{bm}
\usepackage{lipsum}
\usepackage{multirow}
\usepackage{amsmath}
\usepackage{algorithm}
\usepackage{algorithmic}
\usepackage{newfloat}
\usepackage{listings}
\DeclareCaptionStyle{ruled}{labelfont=normalfont,labelsep=colon,strut=off} 
\floatstyle{ruled}
\newfloat{listing}{tb}{lst}{}
\floatname{listing}{Listing}
\title{L4DR: LiDAR-4DRadar Fusion for Weather-Robust 3D Object Detection}
\author{
    Xun Huang\textsuperscript{\rm 1,\rm 2,\rm 3},
    Ziyu Xu\textsuperscript{\rm 1,\rm 2},
    Hai Wu\textsuperscript{\rm 1,\rm 2},
    Jinlong Wang\textsuperscript{\rm 1,\rm 2},
    Qiming Xia\textsuperscript{\rm 1,\rm 2},\\
    Yan Xia\textsuperscript{\rm 4},
    Jonathan Li\textsuperscript{\rm 5},
    Kyle Gao\textsuperscript{\rm 5},
    Chenglu Wen\textsuperscript{\rm 1,\rm 2}\thanks{ Corresponding author.  Email: clwen@xmu.edu.cn.},
    Cheng Wang\textsuperscript{\rm 1,\rm 2}
}
\affiliations{
    \textsuperscript{\rm 1}Fujian Key Laboratory of Sensing and Computing for Smart Cities, Xiamen University, China \\
    \textsuperscript{\rm 2}Key Laboratory of Multimedia Trusted Perception and Efficient Computing, \\Ministry of Education of China, Xiamen University, China \\
    \textsuperscript{\rm 3}Zhongguancun Academy, China  \\
    \textsuperscript{\rm 4}Technische Universität München, Germany \\
    \textsuperscript{\rm 5}University of Waterloo, Canada
}

\begin{document}

\maketitle

\begin{abstract}
LiDAR-based 3D object detection is crucial for autonomous driving. However, due to the quality deterioration of LiDAR point clouds, it suffers from performance degradation in adverse weather conditions.
Fusing LiDAR with the weather-robust 4D radar sensor is expected to solve this problem; however, it faces challenges of significant differences in terms of data quality and the degree of degradation in adverse weather. 
 To address these issues, we introduce L4DR, a weather-robust 3D object detection method that effectively achieves LiDAR and 4D Radar fusion. Our L4DR proposes \textbf{M}ulti-\textbf{M}odal \textbf{E}ncoding (MME) and \textbf{F}oreground-\textbf{A}ware \textbf{D}enoising (FAD) modules to reconcile sensor gaps, which is the first exploration of the complementarity of early fusion between LiDAR and 4D radar. Additionally, we design an \textbf{I}nter-\textbf{M}odal and \textbf{I}ntra-\textbf{M}odal (\{IM\}$^2$) parallel feature extraction backbone coupled with a \textbf{M}ulti-\textbf{S}cale \textbf{G}ated \textbf{F}usion (MSGF) module to counteract the varying degrees of sensor degradation under adverse weather conditions. Experimental evaluation on a VoD dataset with simulated fog proves that L4DR is more adaptable to changing weather conditions. It delivers a 
significant performance increase under different fog levels, improving the 3D mAP by up to 20.0\% over the traditional LiDAR-only approach. Moreover, the results on the K-Radar dataset validate the consistent performance improvement of L4DR in real-world adverse weather conditions. 

\end{abstract}

\begin{links}
     \link{Code}{https://github.com/ylwhxht/L4DR}
     \link{Extended version}{https://arxiv.org/abs/2408.03677}
 \end{links}

\section{Introduction}

3D object detection is a fundamental vision task of unmanned platforms, extensively utilized in applications such as intelligent robot navigation \citep{robots1,robots2} and autonomous driving~\citep{stf}. For example, Full Driving Automation (FDA, Level 5) relies on weather-robust 3D object detection, which provides precise 3D bounding boxes even under various challenging adverse weather conditions~\citep{MVDNet}. Owing to the high resolution and strong interference resistance of LiDAR sensors, LiDAR-based 3D object detection has emerged as a mainstream area of research \citep{cpd,hinted,cmd}. However, LiDAR sensors exhibit considerable sensitivity to weather conditions. In adverse scenarios, the scanned LiDAR point clouds suffer from substantial degradation and increased noise~\citep{SRKD, fogsim}. This degradation negatively impacts 3D detectors, compromising the reliability of autonomous perception systems.

\begin{figure}[!t]

    \includegraphics[width=1.0\columnwidth]{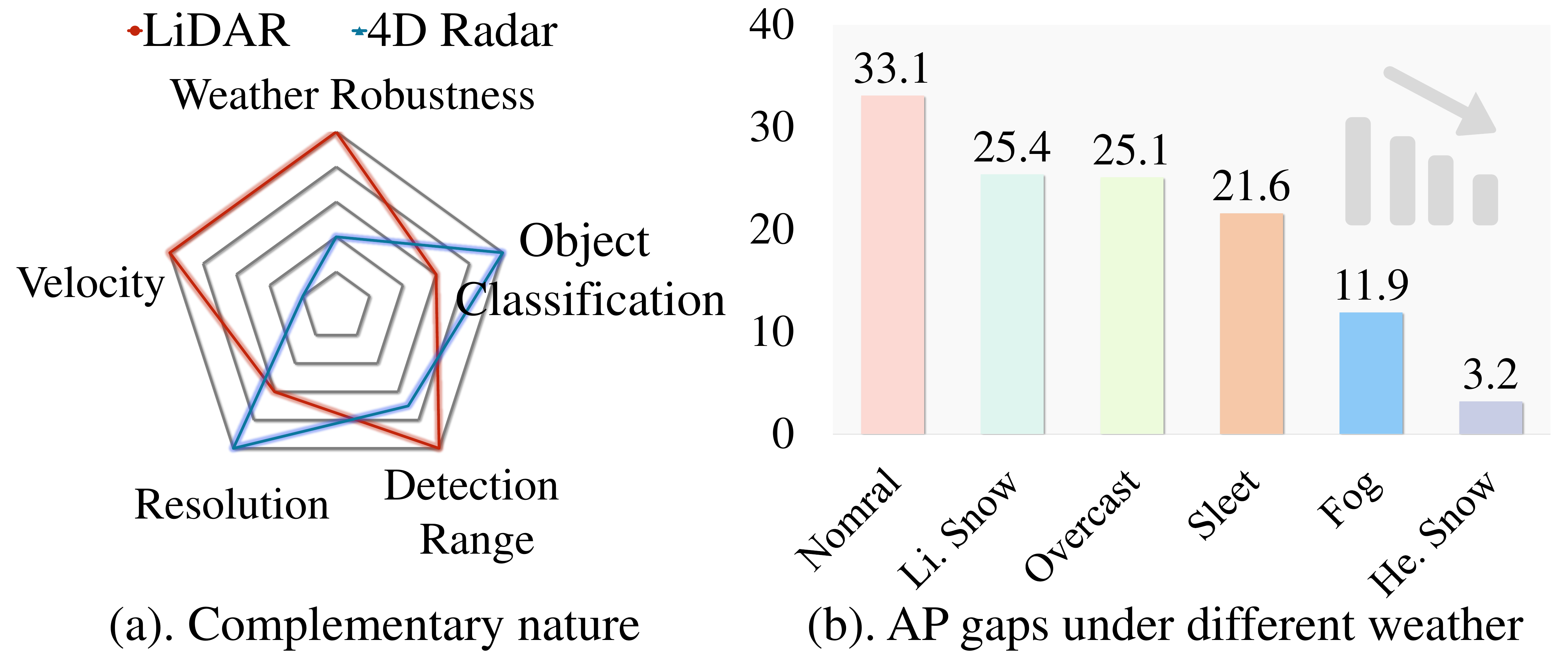}
  \caption{\textcolor{black}{(a) The radar chart illustrates the complementary nature of LiDAR and 4D radar sensors. (b) AP gaps between LiDAR and 4D radar in real different weather (the extent to which LiDAR is superior to 4D radar). }}
  \label{intro}
\end{figure}

\begin{figure*}[!h]
    
    \includegraphics[width=\textwidth]{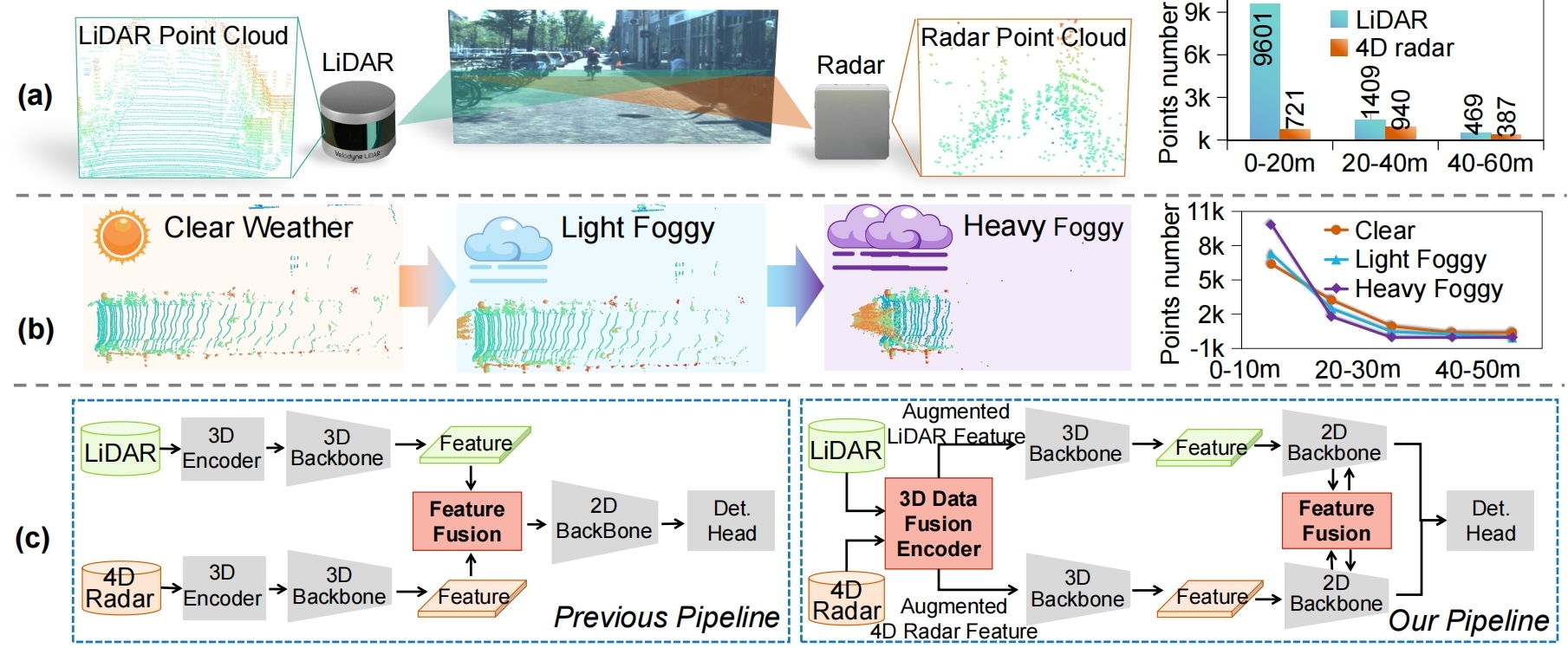}
  \centering
  \caption{(a) Significant quality disparity between the LiDAR and the 4D radar. (b) Severe degradation of LiDAR data quality in adverse weather. (c) Comparison of previous LiDAR-4DRadar fusion and our fusion, highlighting our innovative framework designs to address challenges (a) and (b).}
  \label{challenge}
\end{figure*}
\begin{figure}[!t]

    \includegraphics[width=1.0\columnwidth]{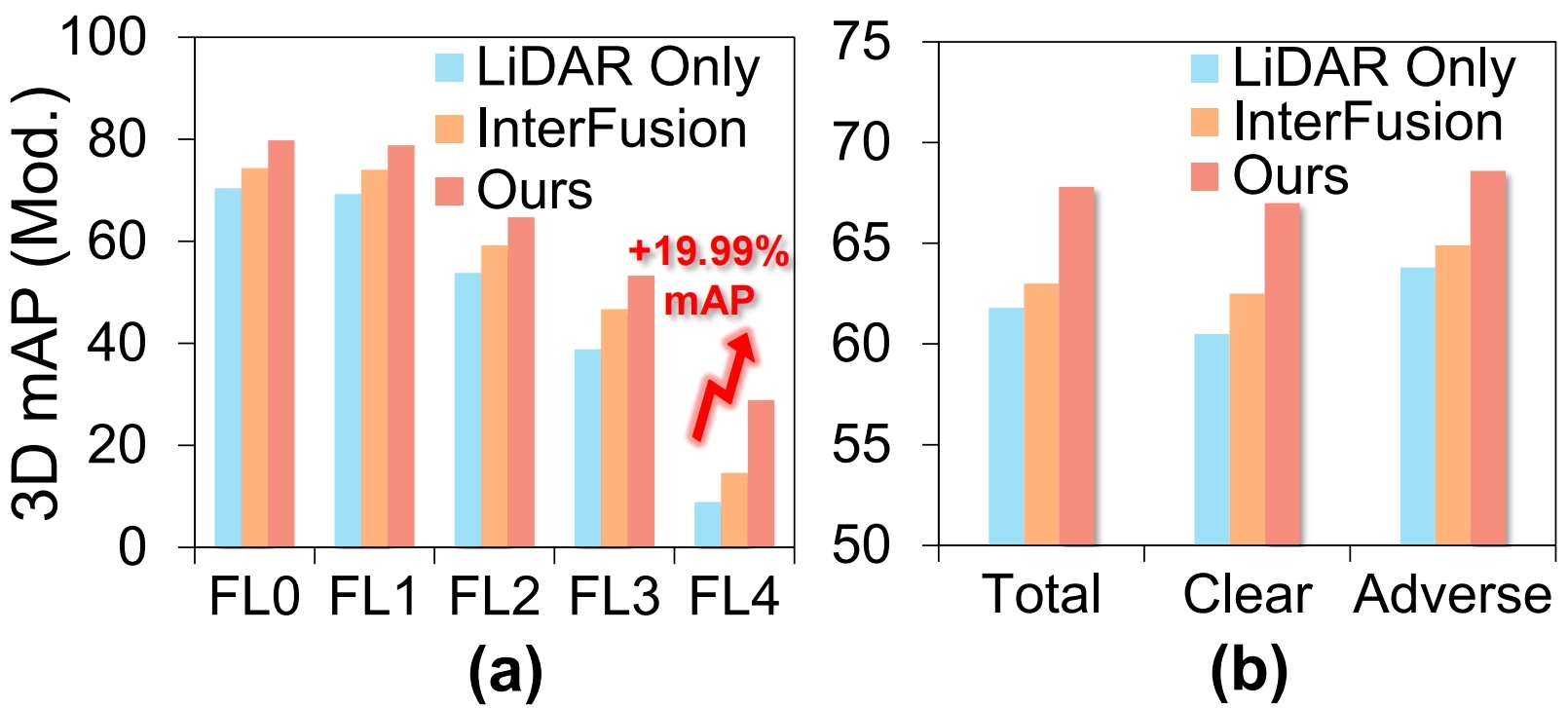}
  \caption{Performance comparison of our L4DR and LiDAR-only in (a) various simulated fog levels (FL denotes fog level) and (b) real-world adverse weather.}
  \label{performance}
\end{figure}

\textcolor{black}{Aside from LiDAR, 4D (range, azimuth, Doppler, and elevation) millimeter-wave radar is gaining recognition as a vital perception sensor \citep{survey_4d1,survey_4d2,lalala}. As shown in Fig.\ref{intro} (a), 4D radar outperforms LiDAR in weather robustness, velocity measurement, and detection range. The millimeter-wave signals of 4D  radar have wavelengths much larger than the tiny particles in fog, rain, and snow~\citep{wavelength1, wavelength2}, exhibiting reduced susceptibility to weather disturbances. As shown in Fig.\ref{intro} (b), the performance gap between LiDAR and 4D radar decreases as the severity of the weather rises. Hence, the 4D radar sensor is more suitable for adverse weather conditions than the LiDAR sensor. However, LiDAR has advantages in terms of higher distance and angular point resolution. These circumstances make it feasible to promote the full fusion of 4D radar and LiDAR data to improve 3D object detection. Pioneering approaches such as InterFusion~\citep{InterFusion}, M$^2$Fusion~\citep{MMFusion}, and 3D-LRF \cite{L4DR} conducted initial attempts to fuse LiDAR and 4D radar data, have demonstrated significant performance improvements over single-sensor models. }

Despite progress, the significant data quality disparities and varying sensor degradation in adverse weather conditions remain largely unaddressed. As shown in Fig.\ref{challenge} (a), a primary challenge arises from the significant quality disparity between the LiDAR and 4D radar sensors.
A second challenge pertains to varying sensor degradation under adverse weather conditions. Fig.\ref{challenge} (b) shows LiDAR sensors undergo severe data degradation in adverse weather. Conversely, the data quality decrease in 4D radar is significantly lower \citep{survey_4d1,survey_4d2} (Details are given in extended version \cite{our}). \textcolor{black}{The varying degradation of sensors leads to fluctuations in the fused features, resulting in difficulty for the model to maintain high performance under adverse weather conditions. Therefore, developing a weather-robust 3D object detector requires overcoming the challenges of significant quality disparities and varying degradation.}

\textcolor{black}{To address the above challenges, we propose L4DR, a novel two-stage fusion framework for LiDAR and 4D radar (shown in Fig. \ref{challenge} (c)). The first-stage fusion is performed by the 3D Data Fusion Encoder, which contains the Foreground-Aware Denoising (FAD) module and the Multi-Modal Encoder (MME) module. This first-stage data fusion augments LiDAR and 4D radar with each other to tackle the challenge of significant data quality disparities between LiDAR and 4D radar. The second-stage Feature Fusion is to fuse features between 2D Backbones to address the second challenge of the varying sensor degradation. It consists of the Inter-Modal and Intra-Modal (\{IM\}$^2$) backbone and the Multi-Scale Gated Fusion (MSGF) module. Different from traditional methods that fuse features before extraction on the left side of Fig.\ref{challenge} (c), L4DR integrates a continuous fusion process throughout the feature extraction, adaptively focusing on the significant modal features in different weather conditions.}

In Fig.~\ref{performance}, our comprehensive experimental results 
showcase the superior performance of L4DR across various simulated and real-world adverse weather disturbances. 
Our main contributions are as follows:

\begin{itemize}[leftmargin=10pt]

\item We introduce the innovative Multi-Modal Encoder (MME) module, which achieves LiDAR and 4D radar data early fusion without resorting to error-prone processes (e.g., depth estimation). It effectively bridges the substantial LiDAR and 4D radar data quality disparities. 

\item We designed an \{IM\}$^2$ backbone with a  Multi-Scale Gated Fusion (MSGF) module, adaptively extracting salient features from LiDAR and 4D radar in different weather conditions. This enables the model to adapt to varying levels of sensor degradation under adverse weather conditions.

\item Extensive experiments on the two benchmarks, VoD and K-Radar, demonstrate the effectiveness of our L4DR under various levels and types of adverse weather, achieving new state-of-the-art performances on both datasets. 

\end{itemize}

\begin{figure*}[!t]
    
    \includegraphics[width=\textwidth]{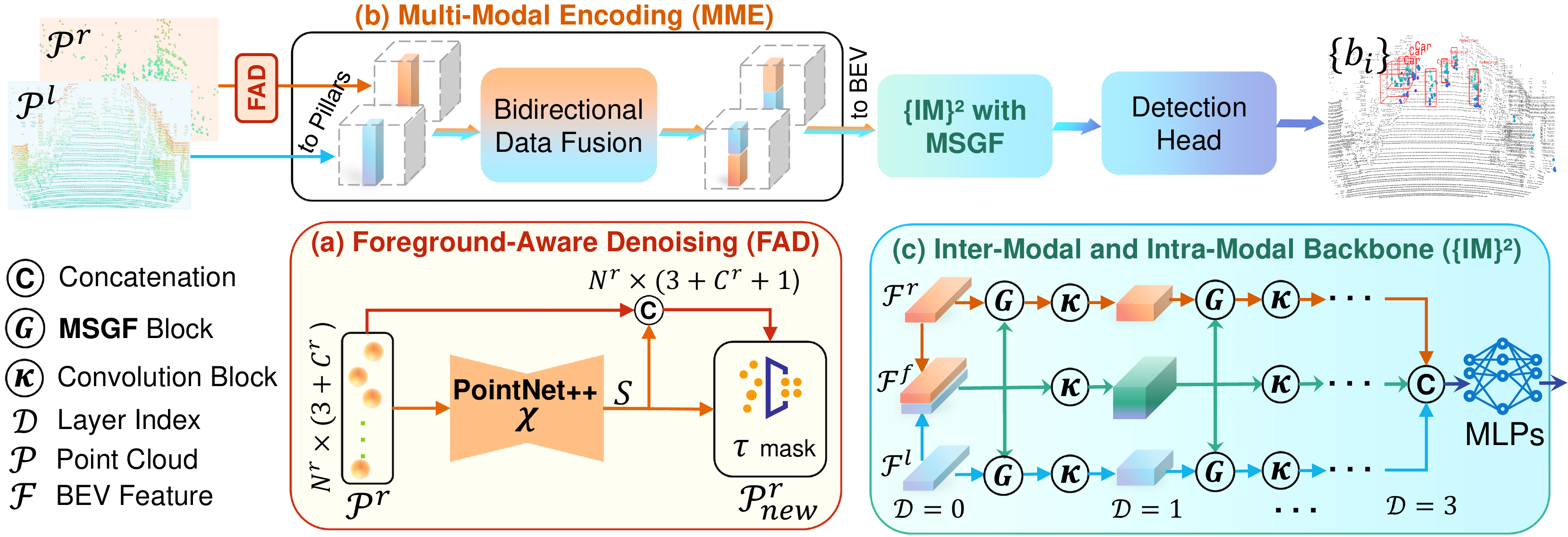}
  \centering
  \caption{L4DR framework. (a) \textbf{F}oreground-\textbf{A}ware \textbf{D}enoising (FAD) performs denoising by segmenting foreground semantics per 4D radar point. Next, (b) \textbf{M}ulti-\textbf{M}odal \textbf{E}ncoder (MME) fuses bi-directional data for both LiDAR and 4D radar modalities at the Encoder stage to obtain higher quality BEV features. Finally, (c) \textbf{I}nter-\textbf{M}odal and \textbf{I}ntra-\textbf{M}odal (\{IM\}$^2$) backbone coupled with \textbf{M}ulti-\textbf{S}cale \textbf{G}ated \textbf{F}usion (MSGF) uses a gating strategy to filter features to avoid redundant information while extracting inter-modal and intra-modal features in parallel.}
  \label{frame}
\end{figure*}




\section{Related Work}

\paragraph{LiDAR-based 3D object detection.}
Researchers have developed single-stage and two-stage methods to tackle challenges for 3D object detection. Single-stage detectors such as VoxelNet~\citep{second}, PointPillars~\citep{PointPillars}, 3DSSD~\citep{3DSSD}, DSVT \citep{dsvt} utilize PointNet++ \citep{pointnet++}, sparse convolution, or other point feature encoder to extract features from point clouds and perform detection in the Bird's Eye View (BEV) space. Conversely, methods such as PV-RCNN~\citep{pvrcnn}, PV-RCNN++~\citep{pvrcnn++}, Voxel-RCNN \citep{voxelrcnn}, and VirConv ~\citep{virconv} focus on two-stage object detection, integrating RCNN networks into 3D detectors. Even though these mainstream methods have gained excellent performance in normal weather, they still lack robustness under various adverse weather conditions.

\paragraph{LiDAR-based 3D object detection in adverse weather.}
LiDAR sensors may undergo degradation under adverse weather conditions. Physics-based weather simulations~\cite{snowsim,fogsim, LISA} have been explored to produce point clouds under adverse weather to alleviate data scarcity. \cite{denoise_snow,cnn_denoise} utilized the DROR, DSOR, or CNN to classify and filter LiDAR noise points. \cite{spg} designed a general completion framework that addresses the domain gaps across different weather conditions. \cite{SRKD} designed a general knowledge distillation framework that transfers sunny performance to rainy performance. However, these methods primarily rely on single-LiDAR modal data, which will be constrained by the decline in the quality of LiDAR under adverse weather conditions.

\paragraph{LiDAR-radar fusion-based 3D object detection.} MVDNet~\citep{MVDNet} has designed a framework for fusing LiDAR and radar. ST-MVDNet~\citep{ST-MVDNet} and ST-MVDNet++~\citep{ST-MVDNet++} incorporate a self-training teacher-student to MVDNet to enhance the model. Bi-LRFusion~\citep{Bi-LRFusion} employs a bidirectional fusion strategy to improve dynamic object detection performance. However, these studies only focus on 3D radar and LiDAR. As research progresses, the newest studies continue to drive the development of LiDAR-4D radar fusion. M$^2$-Fusion~\citep{MMFusion}, InterFusion~\citep{InterFusion}, and 3D-LRF \cite{L4DR} explore LiDAR and 4D radar fusion. However, these methods have not considered and overcome the challenges of fusing 4D radar and LiDAR under adverse weather conditions.

\section{Methodology}

\subsection{Problem Statement and Overall Design}
\paragraph{LiDAR-4D radar fusion-based 3D object detection.} \textcolor{black}{For an outdoor scene, we denote LiDAR point cloud as
$\mathcal{P}^l = \{p^l_i\}_{i=1}^{N_l}$
and 4D radar point cloud as 
$\mathcal{P}^r = \{p^r_i\}_{i=1}^{N_r}$, 
where $p$ denotes 3D points.
Subsequently, a multi-modal model $\mathcal{M}$ extract deep features $\mathcal{F}^m$ from $\mathcal{P}^m$,  written as $\mathcal{F}^m = g(f_\mathcal{M}(\mathcal{P}^m; \Theta))$, where $m\in\{l,r\}$.
The fusion features are then obtained by $\mathcal{F}^f = \phi(\mathcal{F}^l, \mathcal{F}^r)$, where $\phi $ donates fusion method. 
The objective of 3D object detection is to regress the 3D bounding boxes $B = \{b_i\}_{i=1}^{N_b}$, $B \in \mathbb{R}^{N_b\times7}$.}

\paragraph{Significant data quality disparity.} \textcolor{black}{As mentioned before, there is a dramatic difference between $\mathcal{P}^l$ and $\mathcal{P}^r$ in the same scene. To fully fuse two modalities, we can use $P^l$ to enhance the highly sparse $P^r$ that lacks discriminative details. Therefore, our L4DR includes a \textbf{M}ulti-\textbf{M}odal \textbf{E}ncoder (\textbf{MME}, Figure \ref{frame} (b)), which performs data early-fusion complementarity at the encoder. However, we found that direct data fusion would also cause noises in $\mathcal{P}^r$ to spread to $\mathcal{P}^l$. Therefore, we integrated \textbf{F}oreground-\textbf{A}ware \textbf{D}enoising (\textbf{FAD}, Figure \ref{frame} (a)) into L4DR before MME to filter out most of the noise in $\mathcal{P}^r$.}

\paragraph{Varying sensor degradation in adverse weather.}
\textcolor{black}{Compared to 4D radar, the quality of LiDAR point cloud $P^l$ is more easily affected by adverse weather conditions, leading to varying feature presentations $\mathcal{F}^l$. 
Previous backbones focusing solely on fused inter-modal features $\mathcal{F}^f$ overlook the weather robustness of 4D radar, leading to challenges in addressing the frequent fluctuation of $\mathcal{F}^l$. To ensure robust fusion across diverse weather conditions, we introduce the \textbf{I}nter-\textbf{M}odal and \textbf{I}ntra-\textbf{M}odal (\{IM\}$^2$, Figure \ref{frame} (c)) backbone. This design simultaneously focuses on inter-modal and intra-modal features, enhancing model adaptability. However, redundancy between these features arises. Inspired by gated fusion techniques \cite{GF1, GF2}, we propose the \textbf{M}ulti-\textbf{S}cale \textbf{G}ated \textbf{F}usion (MSGF, Figure \ref{frame} (d)) module. MSGF utilizes inter-modal features $\mathcal{F}^f$ to filter intra-modal features $\mathcal{F}^l$ and $\mathcal{F}^r$, effectively reducing feature redundancy.}

\subsection{Foreground-Aware Denoising (FAD)}
Due to multipath effects, 4D radar contains significant noise points. Despite applying the Constant False Alarm Rate (CFAR) algorithm to filter out noise during the data acquisition process, the noise level remains substantial. It is imperative to further reduce clutter noise in 4D radar data before early data fusion to avoid spreading noise. Considering the minimal contribution of background points to object detection, this work introduces point-level foreground semantic segmentation to 4D radar denoising, performing a Foreground-Aware Denoising. Specifically, we first utilize PointNet++ \citep{pointnet++} combined with a segmentation head as $\chi$, to predict the foreground semantic probability $\mathcal{S} = \chi(\mathcal{P}^r)$ for each point in the 4D radar. Subsequently, points with a foreground probability below a predefined threshold $\tau$ are filtered out, that is $\mathcal{P}^r_{new} = \{p^r_i|\mathcal{S}_i \ge \tau \}$. 
FAD effectively filters out as many noise points as possible while preventing the loss of foreground points.

\subsection{Multi-Modal Encoder (MME)}

Even following denoising using FAD, there remains a significant quality disparity between LiDAR and 4D radar due to limitations in resolution. We thus design a Multi-Modal Encoder module that fuses LiDAR and radar points at an early stage to extract richer features.

\begin{figure}[!t]
\centering
    \includegraphics[width=0.8\columnwidth]{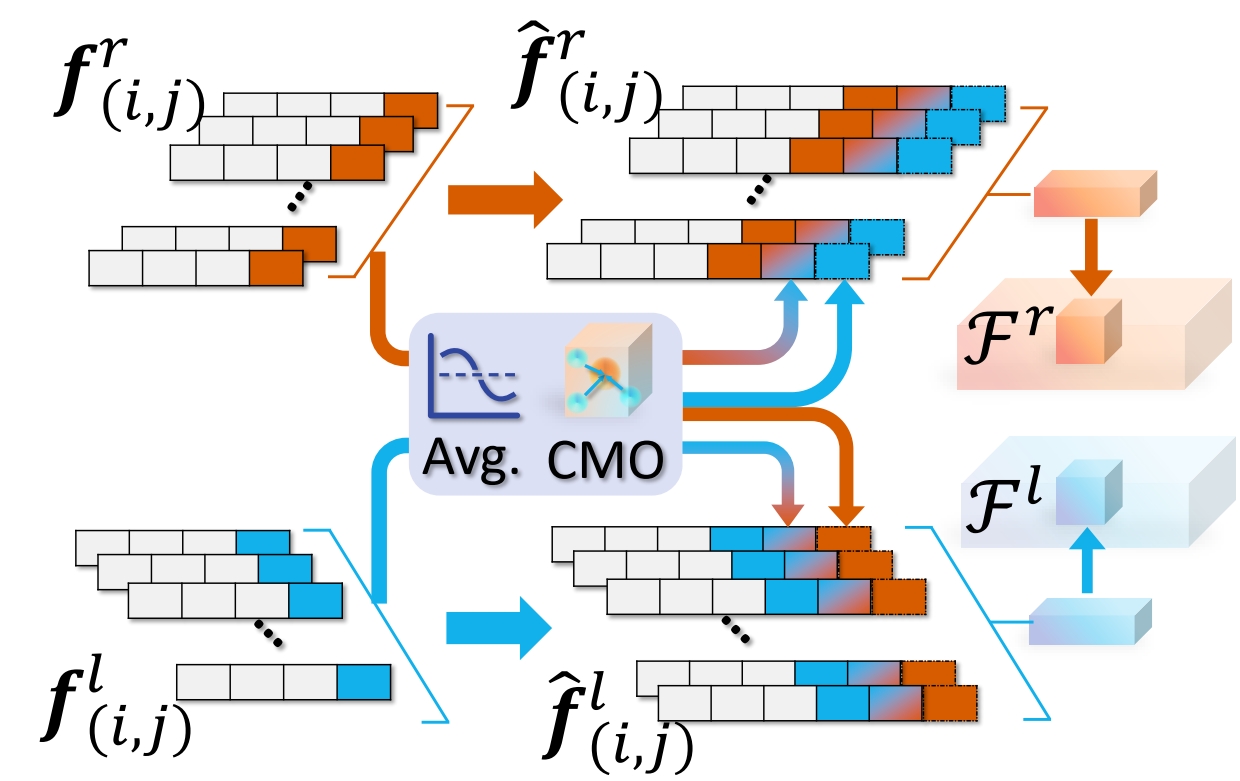}
  \caption{Bidirectional Data Fusion in MME. LiDAR-specific point features (blue) and radar-specific point features (red) located in the same pillar are averaged (Avg.) for feature propagation. And the cross-modal offsets (CMO) are computed to enrich the geometric features.}
  \label{mme}
\end{figure}

As illustrated in Figure \ref{mme}, we innovate the traditional unimodal Pillars coding into multimodal Pillars coding to perform initial fusion at the data level, extracting richer information for subsequent feature processing. Firstly, referring \citep{PointPillars}, we encode the LiDAR point cloud into a pillar set $\mathcal{H}^l=\{h^l_i\}_{i=1}^{N}$. Each LiDAR point $p^l_{(i,j)}$ in pillar $h^l_i$ encoded using encoding feature $\boldsymbol{f}^l_{(i,j)}$ as 
\begin{equation}
\label{eq3}
\boldsymbol{f}^l_{(i,j)}= 
[\boldsymbol{\mathcal{X}}^l,\boldsymbol{\mathcal{Y}}^l_{cl},\boldsymbol{\mathcal{Z}}^l,\lambda],
\end{equation}
where $\boldsymbol{\mathcal{X}}^l=[x^l,y^l,z^l]$ is the coordinate of the LiDAR point,  $\boldsymbol{\mathcal{Y}}^l_{cl}$ denotes the distance from the LiDAR point to the arithmetic mean of all LiDAR points in the pillar,  $\boldsymbol{\mathcal{Z}}^l$ denotes the (horizontal) offset from the pillar center in $x,y$ coordinates, and $\lambda$ is the reflectance. Similarly, each 4D radar point $p^r_{(i,k)}$ in pillar $h^r_i$ encoded using  encoding feature $\boldsymbol{f}^r_{(i,k)}$ as 
\begin{equation}
\boldsymbol{f}^r_{(i,k)}= 
[\boldsymbol{\mathcal{X}}^r,\boldsymbol{\mathcal{Y}}^r_{cr},\boldsymbol{\mathcal{Z}}^r,\boldsymbol{\mathcal{V}},\Omega],
\end{equation}
where $\boldsymbol{\mathcal{X}}^r$, $\boldsymbol{\mathcal{Y}}^r_{cr}$, and $\boldsymbol{\mathcal{Z}}^r$ are similar in meaning to those in Eq.\ref{eq3}. $\boldsymbol{\mathcal{V}}=[\mathcal{V}_r,\mathcal{V}_a]$ is the relative and absolute radial velocity of Doppler information, and $\Omega$ is Radar Cross-Section (RCS).

We then perform cross-modal feature propagation for LiDAR and radar pillar encoding features that occupy the same pillar coordinates. The fused LiDAR and 4D radar pillar encoding features $\widehat{\boldsymbol{f}}^l_{(i,j)}$ and $\widehat{\boldsymbol{f}}^r_{(i,k)}$ are obtained by fusing $\boldsymbol{f}^l_{(i,j)}$ and $\boldsymbol{f}^r_{(i,k)}$ as follows:
\begin{equation}
\begin{aligned}
\widehat{\boldsymbol{f}}^l_{(i,j)}=[\boldsymbol{\mathcal{X}}^l,\boldsymbol{\mathcal{Y}}^l_{cl},\boldsymbol{\mathcal{Y}}^l_{cr},\boldsymbol{\mathcal{Z}}^l,\lambda,\overline{\boldsymbol{\mathcal{V}}},\overline{\Omega}], \\
  \enspace\mathrm{and } \enspace  
 \widehat{\boldsymbol{f}}^r_{(i,k)}=[\boldsymbol{\mathcal{X}}^r,\boldsymbol{\mathcal{Y}}^r_{cl},\boldsymbol{\mathcal{Y}}^r_{cr},\boldsymbol{\mathcal{Z}}^r,\overline{\lambda},\boldsymbol{\mathcal{V}},\Omega],
\end{aligned}
\end{equation}
where the overline denotes the average of all point features of another modality in that pillar.

The feature propagation is beneficial because $\lambda$ and $\Omega$ are helpful for object classification, while Doppler information $\boldsymbol{\mathcal{V}}$ is crucial for distinguishing dynamic objects \citep{LiRaFusion}. Cross-modal feature sharing makes comprehensive use of these advantages and cross-modal offsets [$\boldsymbol{\mathcal{Y}}^m_{cl},\boldsymbol{\mathcal{Y}}^m_{cr}$], $m \in \{l,r\}$ also further enrich the geometric information.
This MME method compensates for the data quality of 4D radar under normal weather conditions and can also enhance the quality of LiDAR in adverse weather. Subsequently,  we applied a linear layer and max pooling operations to the fused pillar encoding features $\widehat{\boldsymbol{f}}$ to obtain the corresponding modal BEV features $\mathcal{F}$.

\begin{figure}
\centering
    \includegraphics[width=0.8\columnwidth]{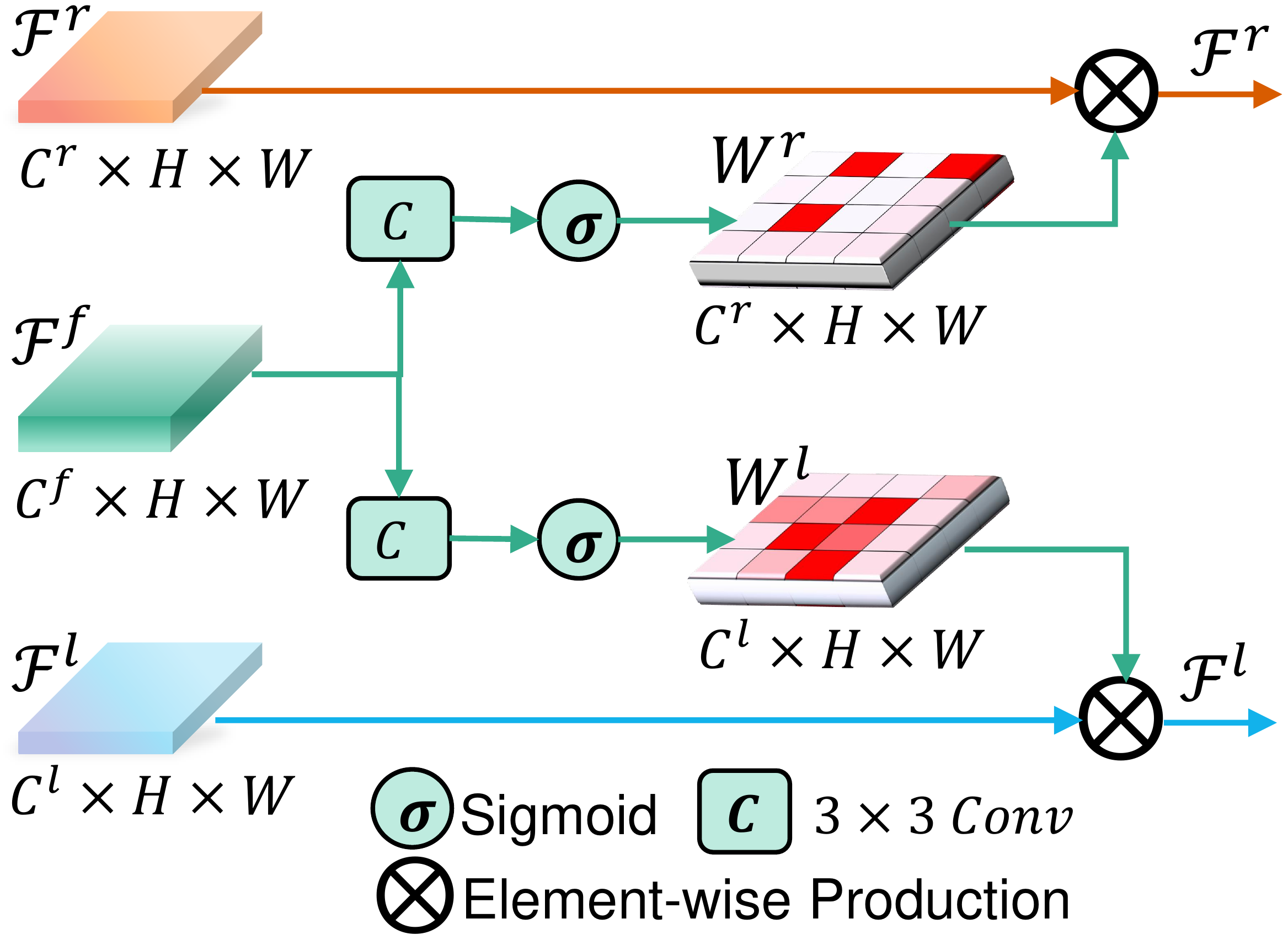}
  \caption{Gated fusion block in our MSGF. The gated block processes input features from LiDAR and 4D radar by using the fused inter-modal features. It generates adaptive gating weights from fused features and applies these weights via element-wise multiplication to filter out redundant information in intra-modal features.}
  \label{gf}
\end{figure}

\subsection{\{IM\}$^2$ Backbone and MSGF Block}

To take full advantage of the respective advantages of LiDAR and 4D radar, it is necessary to focus on both inter-modal features and intra-modal features. We introduce the \textbf{I}nter-\textbf{M}odal and \textbf{I}ntra-\textbf{M}odal backbone (\{IM\}$^2$). \{IM\}$^2$ serves as a multi-modal branch feature extraction module that concurrently extracts inter-modal feature ($\mathcal{F}^f$) and intra-modal features ($\mathcal{F}^l, \mathcal{F}^r$). Specifically, we fuse two Intra-Modal features to form an Inter-Modal feature,
\begin{equation}
\mathcal{F}^f= \phi(\mathcal{F}^l,\mathcal{F}^r),
\end{equation}
where $\phi$ denotes fusion approach (we use concatenation).
Subsequently, we apply a convolutional block to each modal branch $\mathcal{F}^l$, $\mathcal{F}^r$, and $\mathcal{F}^f$ independently,
\begin{equation}
    \mathcal{F}^m_\mathcal{D}=\kappa(\mathcal{F}^m_{\mathcal{D}-1}),
\end{equation}
where $\mathcal{D} \in [1,3]$ denotes layer, and $m \in \{l,r,f\}$ indicates different modality. $\kappa$ represents a convolutional layer with batch normalization and ReLU activation.

However, while \{IM\}$^2$ addresses some deficiencies in feature representation, this naive approach inevitably introduces redundant features. Inspired by \citep{LiRaFusion} to adaptively filter each modal feature, we design an MSGF for adaptive gated fusion on each LiDAR and 4D radar scale feature map.

As depicted in Fig. \ref{gf}, the gated network $\mathcal{G}$ in MSGF processes input feature maps from LiDAR $\mathcal{F}^l$, 4D radar $\mathcal{F}^r$ and fused counterpart $\mathcal{F}^f$. Subsequently, on the LiDAR and 4D radar branches, the adaptive gating weights $\mathcal{W}^l$ for $\mathcal{F}^l$ and $\mathcal{W}^r$ for $\mathcal{F}^r$ were obtained by a convolution block and sigmoid activation function, respectively. These weights are applied to the initial feature via element-wise multiplication, thus enabling filter $\mathcal{F}^l$ and $\mathcal{F}^r$ in the gated mechanism.
Formally, the gated network $\mathcal{G}$ guides $\mathcal{F}^l$ and $\mathcal{F}^l$ at convolution layer index $\mathcal{D}$ to filter out redundant information as following: 
\begin{equation}
\begin{aligned}
    \mathcal{F}^m_\mathcal{D}=\mathcal{G}_\mathcal{D}(\mathcal{F}^m_\mathcal{D},\mathcal{F}^f_\mathcal{D}), m\in\{l,r\},
    \\ \mathrm{and} \enspace \mathcal{G}^m_\mathcal{D}(\mathcal{F}_\mathcal{D}^m,\mathcal{F}_\mathcal{D}^f) = \mathcal{F}_\mathcal{D}^m * \delta(\kappa(\mathcal{F}_\mathcal{D}^f)),
\end{aligned}
\end{equation}
where $\kappa$ is a 3x3 convolution block and $\delta$ a sigmoid function.
$\mathcal{F}^f$ is the fused feature with information about the interactions between modalities. It discerns whether features in $\mathcal{F}^l$ and $\mathcal{F}^r$ are helpful or redundant. Using $\mathcal{F}^f$ for gated filtering can flexibly weight and extract features from $\mathcal{F}^l$ and $\mathcal{F}^r$ while significantly reducing feature redundancy.

\subsection{Loss Function and Training Strategy} \textcolor{black}{
We trained our L4DR with the following losses :
\begin{equation}
\label{eq_all}
\mathcal{L}_{all} = \beta_{cls}\mathcal{L}_{cls}+\beta_{loc} \mathcal{L}_{loc}+\beta_{fad} \mathcal{L}_{fad},
\end{equation}
where $\beta_{cls}$ = 1, $\beta_{loc}$ = 2, $\beta_{fad}$ = 0.5, the $\mathcal{L}_{cls}$ is object classification focal loss, the $\mathcal{L}_{loc}$ is object localization regression loss, and the $\mathcal{L}_{fad}$ is focal classification loss of 4D radar foreground point  in FAD module.
We use Adam optimizer with lr = 1e-3, $\beta_1$ = 0.9, $\beta_2$ = 0.999. }

\section{Experiments}

\begin{table*}[!t]

\centering
    \setlength{\tabcolsep}{3.pt}

\begin{tabular}{cc|ccccccccc}
\hline
{{Methods}}  & {{Modality}}     & {{Metric}} & {{Total}} & {{Normal}} & {{Overcast}} & {{Fog}} & {{Rain}} & {{Sleet}} & {{Lightsnow}} & {{Heavysnow}} \\ \hline\hline
\multirow{2}{*}{\shortstack{RTNH\\(NeurIPS 2022)}}   & \multirow{2}{*}{4DR}
  & $AP_{BEV}$& 41.1  & 41.0  & 44.6 & 45.4 & 32.9 & 50.6 & 81.5  & 56.3  \\
& & $AP_{3D}$ &37.4 &37.6 &42.0 &41.2 &29.2 &49.1 &63.9 &43.1  \\ \hline
\multirow{2}{*}{\shortstack{PointPillars\\(CVPR 2019)}}   & \multirow{2}{*}{L}
  & $AP_{BEV}$ &49.1 &48.2 &53.0 &45.4 &44.2 &45.9 &74.5 &53.8 \\
& & $AP_{3D}$ &22.4 &21.8 &28.0 &28.2 &27.2 &22.6 &23.2 &12.9  \\ \hline
\multirow{2}{*}{\shortstack{RTNH\\(NeurIPS 2022)}}   & \multirow{2}{*}{L}
  & $AP_{BEV}$ &66.3 &65.4 &87.4 &83.8 &73.7 &\underline{48.8} &78.5 &48.1 \\
&  & $AP_{3D}$ &37.8 &39.8 &46.3 &\underline{59.8} &28.2 &\underline{31.4} &50.7 &24.6  \\  \hline
\multirow{2}{*}{\shortstack{InterFusion\\(IROS 2023)}}   & \multirow{2}{*}{L+4DR}
  & $AP_{BEV}$ &52.9 &50.0 &59.0 &80.3 &50.0 &22.7 &72.2 &53.3 \\
&  & $AP_{3D}$ &17.5 &15.3 &20.5 &47.6 &12.9 &9.33 &\underline{56.8} &25.7  \\ \hline
\multirow{2}{*}{\shortstack{3D-LRF\\(CVPR 2024)}}   & \multirow{2}{*}{L+4DR}
  & $AP_{BEV}$ &\underline{73.6} &\underline{72.3} &\underline{88.4} &\underline{86.6} &\underline{76.6} &47.5 &\underline{79.6} &\textbf{64.1} \\
&  & $AP_{3D}$ &\underline{45.2} &\underline{45.3} &\underline{55.8} &51.8 &\underline{38.3} &23.4 &\textbf{60.2} &\underline{36.9}  \\ \hline
\multirow{2}{*}{\shortstack{L4DR\\(Ours)}}   & \multirow{2}{*}{L+4DR}
  & $AP_{BEV}$ &\textbf{77.5} &\textbf{76.8} &\textbf{88.6} &\textbf{89.7} &\textbf{78.2} &\textbf{59.3} &\textbf{80.9} &\underline{53.8} \\
& & $AP_{3D}$  &\textbf{53.5} &\textbf{53.0} &\textbf{64.1} &\textbf{73.2} &\textbf{53.8} &\textbf{46.2} &52.4 &\textbf{37.0}  \\ \hline
\end{tabular}
\caption{\textcolor{black}{Quantitative results of different 3D object detection methods on K-Radar dataset. We present the modality of each method (L: LiDAR, 4DR: 4D radar) and detailed performance for each weather condition. Best in \textbf{bold}, second in \underline{underline}.} }
\label{tab:kradar}

 \end{table*}

\begin{table*}[!t]

\centering
\small
\begin{tabular}{ccc|ccc|ccc|ccc}
\hline
\multirow{2}{*}{Fog Level}   & \multirow{2}{*}{Methods} &\multirow{2}{*}{{Modality}} & \multicolumn{3}{c}{Car (IoU = 0.5)}           & \multicolumn{3}{|c|}{{Pedestrian (IoU = 0.25)}}     & \multicolumn{3}{c}{{Cyclist (IoU = 0.25)}}         \\

&  & & \multicolumn{1}{c}{{Easy}} & \multicolumn{1}{c}{{Mod.}} & {Hard} 
&\multicolumn{1}{c}{{Easy}} & \multicolumn{1}{c}{{Mod.}} & {Hard}  & \multicolumn{1}{c}{{{Easy}}} & \multicolumn{1}{c}{{Mod.}} & {Hard}  \\ \hline\hline
\multirow{3}{*}{\begin{tabular}[c]{@{}c@{}}

0\\ (W/o Fog)\end{tabular}} 

& PointPillars  & L & 84.9        & 73.5        & 67.5          & 62.7        & 58.4        & 53.4          & 85.5        & 79.0        & 72.7 \\

& InterFusion     & L+4DR         & 67.6        & 65.8        & 58.8          & 73.7        & 70.1        & 64.7          & 90.3        & 87.0        & 81.2          \\ 

& L4DR (Ours) & L+4DR & \textbf{85.0}     & \textbf{76.6}     & \textbf{69.4} & \textbf{74.4}     & \textbf{72.3}     & \textbf{65.7} & \textbf{93.4}     & \textbf{90.4}     & \textbf{83.0} \\ \hline
\multirow{3}{*}{1}  
& PointPillars  & L  & 79.9        & 72.7        & 67.0          & 59.9        & 55.6        & 50.5          & 85.5        & 78.2        & 72.0     \\
& InterFusion     & L+4DR          & 66.1        & 64.0        & 56.9          & 74.0        & 70.6        & 64.5          & 91.6        & 87.4        & 82.0          \\ 
& L4DR (Ours)  & L+4DR  & \textbf{77.9}     & \textbf{73.2}     & \textbf{67.8} & \textbf{75.4}     & \textbf{72.1}     & \textbf{66.7} & \textbf{93.8}     & \textbf{91.0}     & \textbf{83.2} \\  \hline
\multirow{3}{*}{2}
& PointPillars  & L  & 67.0        & 51.4        & 44.4          & 53.1        & 47.2        & 42.7          & 69.6        & 62.7        & 57.2 \\
& InterFusion    & L+4DR          & 56.0        & 48.5        & 41.5          & \textbf{63.2}     & 57.8        & 52.9          & 77.3        & 71.1       & 66.2          \\ 
& L4DR (Ours) & L+4DR  & \textbf{68.5}     & \textbf{56.4}     & \textbf{49.3} & 63.1        & \textbf{59.9}     & \textbf{55.1} & \textbf{82.7}     & \textbf{70.8}     & \textbf{70.7} \\ \hline

\multirow{3}{*}{3}  
& PointPillars  & L   & 44.5        & 31.9        & 27.0          & 40.2        & 37.7        & 34.0          & 53.2        & 46.7        & 41.8          \\
& InterFusion    & L+4DR          & 41.2        & 33.1        & 27.0          & 52.9        & 49.2        & 44.8          & 59.9        & 57.7        & 53.1          \\ 
& L4DR (Ours)  & L+4DR & \textbf{46.2}     & \textbf{41.4}     & \textbf{34.6} & \textbf{53.5}     & \textbf{50.6}     & \textbf{46.2} & \textbf{72.2}     & \textbf{67.7}     & \textbf{60.9} \\ \hline
\multirow{3}{*}{4} 
& PointPillars  & L  & 13.0        & 8.77         & 7.19           & 10.6        & 12.9        & 11.3          & 6.15         & 4.89         & 4.57 \\
& InterFusion      & L+4DR          & 15.2        & 10.8        & 8.40           & 25.7        & 25.1        & 22.6          & 6.68         & 7.95         & 6.99           \\ 
& L4DR (Ours)  & L+4DR  & \textbf{26.9}     & \textbf{26.2}     & \textbf{21.6} & \textbf{33.1}     & \textbf{30.7}     & \textbf{27.9} & \textbf{30.3}     & \textbf{29.7}     & \textbf{26.3} \\ \hline
\end{tabular}
\caption{Quantitative results of different methods on Vod-Fog dataset with the KITTI metric under various fog levels.}
\label{tab:vod_kitti}
\end{table*}

\subsection{Implement Details} 
We implement L4DR with PointPillars \citep{PointPillars}, the most commonly used base architecture in radar-based, LiDAR and 4D radar fusion-based 3D object detection. This can effectively verify the effectiveness of our L4DR and avoid unfair comparisons caused by inherent improvements in the base architecture. We set $\tau$ in section 3.2 as 0.3 while training and 0.2 while inferring. We conduct all experiments with a batch size of 16 on 2 RTX 3090 GPUs. Other parameter settings refer to the default official configuration in the OpenPCDet \citep{openpcdet} tool.

\subsection{Dataset and Evaluation Metrics}

\paragraph{K-Radar dataset.} The K-Radar dataset \citep{K-radar} contains 58 sequences with 34944 frames of 64-line LiDAR, camera, and 4D radar data in various weather conditions. According to the official K-Radar split, we used 17458 frames for training and  17536 frames for testing. We adopt two evaluation metrics for 3D object detection: $AP_{3D}$ and $AP_{BEV}$ of the class "Sedan" at IoU = 0.5.

\paragraph{View-of-Delft (VoD) dataset.} The VoD dataset \citep{vod} contains 8693 frames of 64-line LiDAR, camera, and 4D radar data. Following the official partition, we divided the dataset into a training and validation set with 5139 and 1296 frames. All of the methods used the official radar with 5 scans accumulation and single frame LiDAR. Meanwhile, to explore the performance under different fog intensities, following a series of previous work \citep{MVDNet, ST-MVDNet}, we similarly performed fog simulations \citep{fogsim} (with fog level $\mathcal{L}$ from 0 to 4, fog density ($\alpha$) = [0.00, 0.03, 0.06, 0.10, 0.20]) on the VoD dataset and \textcolor{black}{kept 4D Radar unchanged to simulate various weather conditions}. We named it the \textbf{Vod-Fog} dataset in the following. \textcolor{black}{\textit{Noteworthy, we used two evaluation metrics groups on the VoD dataset. The VoD official metrics are used to compare with previous state-of-the-art methods. The KITTI official metrics are used to analyze the performance of objects with different difficulties.}}

\subsection{Results on K-Radar Adverse Weather Dataset}

Following 3D-LRF \cite{L4DR}, we compare our L4DR with different modality-based 3D object detection methods: PointPillars \cite{PointPillars}, RTNH \cite{RTNH}, InterFusion \cite{InterFusion} and 3D-LRF \cite{L4DR}. The results in Table \ref{tab:kradar} highlight the superior performance of our L4DR model on the K-Radar dataset. Our L4DR model surpasses 3D-LRF by 8.3\% in total $AP_{3D}$. This demonstrates that compared to previous fusion, our method utilizes the advantages of LiDAR and 4D Radar more effectively. \textcolor{black}{We observed that all methods perform better in many adverse weather conditions (e.g., Overcast, Fog, etc.) than in normal weather. A possible reason is the distribution differences of labeled objects in different weather conditions. We discuss this counter-intuitive phenomenon in the extended version and other valuable results, such as different IoU thresholds and version labels.}

\subsection{Results on Vod-Fog Simulated Dataset}
We evaluated our L4DR model in comparison with LiDAR and 4D radar fusion methods using the Vod-Fog dataset using the KITTI metrics across varying levels of fog. Table \ref{tab:vod_kitti} demonstrates that our L4DR model outperforms LiDAR-only PointPillars in different difficulty categories and fog intensities. Particularly in the most severe fog conditions (fog level = 4), our L4DR model achieves performance improvements of 17.43\%, 17.8\%, and 24.81\% mAP in moderate difficulty categories, surpassing the gains obtained by InterFusion. Furthermore, our approach consistently exhibits superior performance compared to InterFusion across various scenarios, showcasing the adaptability of our L4DR fusion under adverse weather conditions.

\begin{table}[!b]

\resizebox{\columnwidth}{!}{
\begin{tabular}{cc|ccc|ccc}
\hline
\multirow{2}{*}{{Methods}} & \makebox[0.03\textwidth][c]{\multirow{2}{*}{{Modality}}} & \multicolumn{3}{c|}{{Entire Area}}                 & \multicolumn{3}{c}{{Driving Area}}                 \\  
                                 &                                    & {Car} & {Ped.} & {Cyc.}    & {Car} & {Ped.} & {Cyc.}   \\ \hline
Pointpillars          & 4DR                                & 39.7                             & 31.0                              & 65.1                                        & 71.6                             & 40.5                              & 87.8                                       \\
LXL                    & 4DR                                & 32.8                             & 39.7                              & 68.1                                     & 70.3                             & 47.3                              & 87.9                                   \\ 
FUTR3D               & C+4DR                              & 46.0                             & 35.1                              & 66.0                                       & 78.7                             & 43.1                              & 86.2                                     \\
BEVFusion             & C+4DR                              & 37.9                             & 41.0                              & 69.0                                       & 70.2                             & 45.9                              & 89.5                                    \\
RCFusion        & C+4DR                              & 41.7                             & 39.0                             & 68.3                                        & 71.9                             & 47.5                              & 88.3                                      \\
LXL                   & C+4DR                              & 42.3                             & 49.5                              & 77.1                                        & 72.2                             & 58.3                              & 88.3                                      \\ 
Pointpillars         & L                                  & 66.0                             & 55.6                              & 75.0                                      & 88.7                             & 68.4                             & 88.4                                      \\ 
InterFusion         & L+4DR                              & 66.5                             & 64.5                              & 78.5                                      & 90.7                             & 72.0                              & 88.7                                      \\ 
L4DR (Ours)                             & L+4DR                              & \textbf{69.1}                    & \textbf{66.2}                     & \textbf{82.8}                      & \textbf{90.8}                    & \textbf{76.1}                     & \textbf{95.5}                    \\ \hline
\end{tabular}}
\caption{Comparing L4DR with different state-of-the-art methods of different modalities on VoD dataset with the VoD metrics. The best performances are highlighted in \textbf{bold}.}
\label{tab:vod_vod}
\end{table}

\subsection{Results on VoD Dataset}
 We compared our L4DR fusion performance with different state-of-the-art methods of different modalities on the VoD dataset with \textit{VoD metric}. As shown in Table \ref{tab:vod_vod}, our L4DR fusion outperforms the existing LiDAR and 4D radar fusion method InterFusion \citep{InterFusion} in all categories. We outperformed by 6.8\% in the Cyc. class in the Driving Area. Meanwhile, our L4DR also significantly outperforms other modality-based state-of-the-art methods such as LXL \citep{LXL}. These experimental results demonstrate that our method can comprehensively fuse the two modalities of LiDAR and 4D radar. As a consequence, our L4DR method also shows superior performance even in clear weather.

\subsection{Ablation study}
\paragraph{Effect of each component.} 
We systematically evaluated each component, with the results summarized in Table~\ref{tab:ablation}. The $1^{st}$ row represents the performance of the LiDAR-only baseline model. \textcolor{black}{Subsequent  $2^{nd}$ row and $3^{rd}$ row are fused by directly concatenating the BEV features from LiDAR and 4D radar modality. The results showing the enhancements were observed with the addition of MME and FAD respectively, highlighting that our fusion method fully utilizes the weather robustness of the 4D radar while excellently handling the noise problem of the 4D radar.}  The $4^{th}$ row indicates that the performance boost from incorporating the \{IM\}$^2$ model alone was not substantial, primarily due to feature redundancy introduced by the \{IM\}$^2$ backbone. This issue was effectively addressed by utilizing the MSGF module in the $5^{th}$ row, leading to the most optimal performance.

\begin{table}[!t]

\resizebox{\columnwidth}{!}{

\begin{tabular}{cccc|ccccc}

\hline
\multicolumn{4}{c|}{Module}                                                           & \multicolumn{5}{c}{3D mAP}\\
\makebox[0.03\textwidth][c]{MME} & \makebox[0.03\textwidth][c]{FAD} & \makebox[0.03\textwidth][c]{\{IM\}$^2$}  & \makebox[0.04\textwidth][c]{MSGF} &  W/o Fog     &    $\mathcal{L}$ = 1       &    $\mathcal{L}$ = 2       &   $\mathcal{L}$ = 3        &  $\mathcal{L}$ = 4         \\ \hline

\Large
                         &                         &                           &      & 70.3     & 68.9     & 53.8     & 38.8     & 8.92     \\
\checkmark                        &                         &                           &      & 77.1     & 75.4     & 63.2     & 52.3     & 23.4     \\
\checkmark                        & \checkmark                       &                           &      & 78.7     & 77.6     & 63.3     & 52.2     & 24.7     \\
\checkmark                        & \checkmark                       & \checkmark                         &      & 78.1     & 76.8     & 62.0     & 52.3     & 26.5     \\
\checkmark                        & \checkmark                       & \checkmark                         & \checkmark    & \textbf{79.8}  & \textbf{78.8}  & \textbf{64.7}  & \textbf{53.3}  & \textbf{28.9}  \\ \hline
\end{tabular}}
\caption{Effect of each component proposed in L4DR, tested on Vod-Fog split. 3D mAP represents the average 3D AP of all classes in Mod. difficulty. $\mathcal{L}$ denotes fog level. }
\label{tab:ablation}
\end{table}

\begin{table}[!b]

\resizebox{\columnwidth}{!}{
\begin{tabular}{c|ccccc}
\hline
\multirow{2}{*}{Fusion} & \multicolumn{5}{c}{3D mAP}\\  &  W/o Fog     &    $\mathcal{L}$ = 1       &    $\mathcal{L}$ = 2       &   $\mathcal{L}$ = 3        &  $\mathcal{L}$ = 4         \\ \hline
Concat.           & 77.9                                                            & 76.3                                                            & 61.9                                                            & 49.3                                                            & 17.7                                                            \\
Cross-Attn.    & 77.2                                                            & 76.0                                                            & 63.0                                                            & 52.7                                                            & \textbf{30.1}                                                   \\
Self-Attn.            & 78.4                                                            & 77.4                                                            & 64.3                                                            & 52.8                                                            & 25.8                                                            \\
SE Block                    & 77.3                                                            & 77.9                                                            & 63.8                                                            & 50.1                                                            & 25.0                                                            \\
CBAM Block                     & 78.0                                                            & 78.1                                                            & 64.0                                                            & 52.3                                                            & 26.4                                                            \\
MSGF (Ours)                     & \textbf{79.8}                                                   & \textbf{78.8}                                                   & \textbf{64.7}                                                   & \textbf{53.2}                                                   & 28.8                                                           \\ \hline
\end{tabular}}
\caption{Comparison of multi-modal feature fusion methods on Vod-Fog val split. 3D mAP represents the average 3D AP of all classes in Mod. difficulty. $\mathcal{L}$ denotes fog level.   }

    \label{tab:fusion}
\end{table}

\paragraph{Comparison with other feature fusion.} We compared different multi-modal feature fusion blocks, including basic concatenation (Concat.) and various attention-based methods such as Transformer-based \citep{transfomer} Cross-Modal Attention (Cross-Attn.) and Self-Attention (Self-Attn.), SE Block \citep{senet}, and CBAM Block \citep{cbam}, see extended version for detailed fusion implements. Experimental results (Table \ref{tab:fusion}) show that while attention mechanisms outperform concatenation to some extent, they do not effectively address the challenge of fluctuating features under varying weather conditions. In contrast, our proposed MSGF method, focusing on significant features of LiDAR and 4D radar, achieves superior performance and robustness under different weather.

\begingroup



\paragraph{Foreground semantic segmentation results in FAD}
To explore the 4D radar denoising ability of FAD, we tested different $\lambda$ in the inference stage, as shown in Table \ref{denoise}. With hyperparameter $\lambda$ increase, recall gradually decreases, and denoise rate, mIoU, and PA gradually increase. We find that the performance of the 3D object detection is not excellent when $\lambda$ is relatively high (0.5). The reason is missing foreground points is more unacceptable than retaining some noise for object detection. Additionally, we also discuss the effect of $\lambda$ on the detection performance in the extended version.
\begin{table}[!t]
 \centering

    \begin{tabular}{ccccc}
    \toprule
    $\lambda$ in testing  & Denoise Rate & Recall & mIoU & PA     \\ \midrule
     
    0.1                            & 88.76        & 84.35                          & 38.07                       & 88.43 \\ 
     
    0.2                              & 94.68       & 78.04                          & 50.11                       & 93.45 \\ 
     
    0.3                               & 96.51      & 73.15                          & 54.14                       & 94.78 \\ 
     
    0.5                               & 97.92      & 65.52                          & 54.39                       & 95.52 \\ \bottomrule
    \end{tabular}
    \caption{Denoise rate, recall, mIoU, and PA (Point Accuracy) of foreground segmentation in FAD ($\lambda $  in training is 0.3).}
    \label{denoise}
     
\end{table}

\section{Conclusion}
In this paper, we analyzed the challenges of fusing LiDAR and 4D radar in adverse weather and proposed L4DR, an effective LiDAR and 4D radar fusion method. We provide an innovative and feasible solution for achieving weather-robust outdoor 3D object detection in various weather conditions. Our experiments on VoD and K-Radar datasets have demonstrated the effectiveness and superiority of our method in various simulated fog levels and real-world adverse weather. In summary, our proposed L4DR paves the way for enhanced reliability in autonomous driving and other applications.  

\noindent\textbf{Limitations.} While the introduction of the \{IM\}$^2$ and MSGF modules has allowed the model to focus on more salient features, it inevitably introduces additional computations that reduce the computational efficiency to a certain extent. The inference speed is reduced to about \textbf{10 FPS}, which is just enough to satisfy the real-time threshold (equal to the LiDAR acquisition frequency), but the computational performance optimization is a valuable future area of research. 

\section{Acknowledgements} This work was supported in part by the National Natural Science Foundation of China (No.62171393), and the Fundamental Research Funds for the Central Universities (No.20720220064).

\section{Appendix / Supplemental Material}

\subsection{Analysis of Point Distribution of LiDAR and 4D Radar under Different Weather Conditions}
Although the weather robustness advantage of 4D radar sensors has been mentioned as a priori knowledge in existing work \citep{survey_4d1, survey_4d2}, this aspect remains less studied. Here, we utilize a variety of real-world adverse weather datasets from K-Radar to examine and corroborate this phenomenon. As depicted in Figure \ref{dist}, we have compiled plots of the point counts averaged across various types of real-world adverse weather conditions for both LiDAR and 4D radar. It is observed that under different categories of adverse weather conditions, the point counts of LiDAR at different distances from the sensor location (a) exhibit a pronounced decreasing trend, reflecting the significant degradation of LiDAR data quality in adverse weather. In contrast, the point counts of 4D radar at different distances from the sensor location (b) do not show a clear correlation with weather conditions. It is important to note that the large differences in data scenes and dynamic object distributions, and the sensitivity of 4D radar to dynamic objects result in greater fluctuations in point count distribution. However, the lack of correlation between point counts and weather conditions still demonstrates to a certain extent the weather robustness advantage of 4D radar.

\begin{figure}[!h]
    \includegraphics[width=\columnwidth]{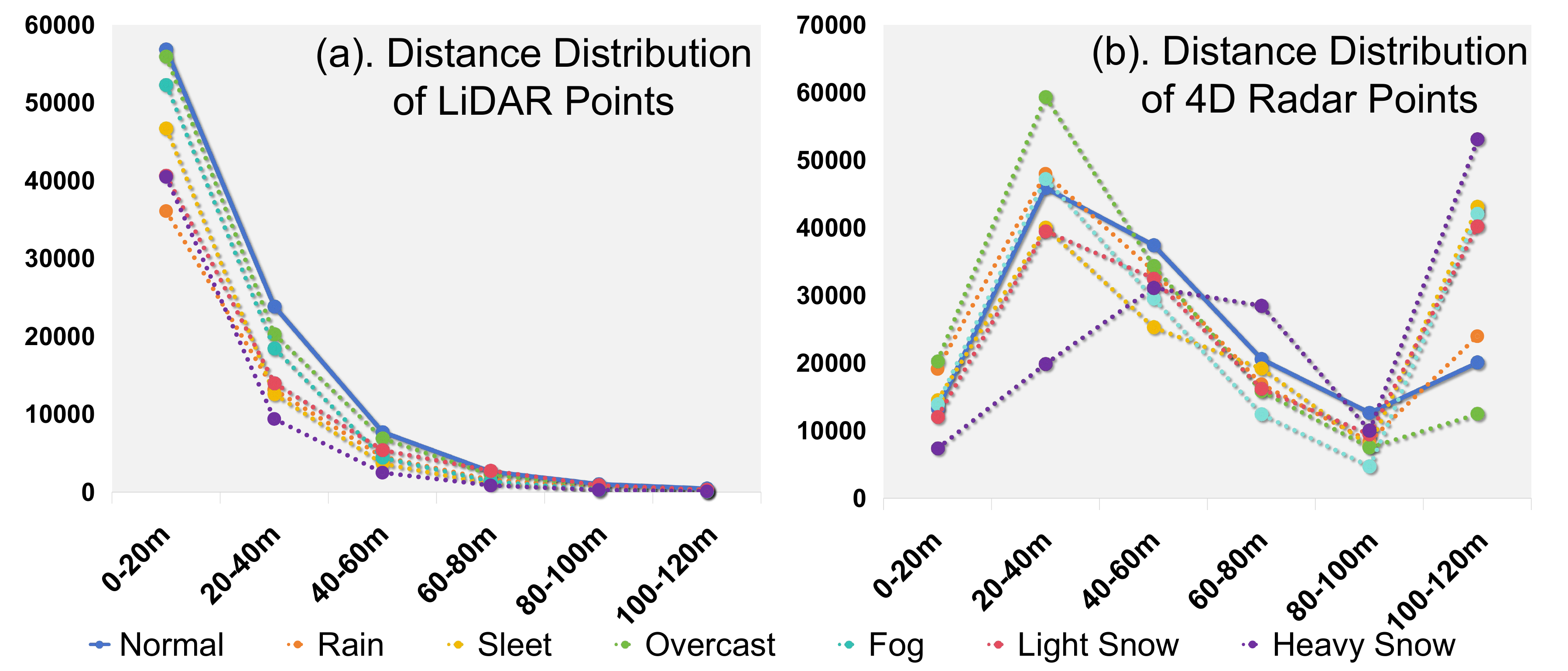}
  \centering
  \caption{Average number of point distribution of (a) LiDAR and (b) 4D radar with distance under different weather conditions (Normal, Rain, Sleet, Overcast, Fog, Light Snow, and Heavy Snow).}
  \label{dist}
\end{figure}

\begin{table*}[!t]
\centering
\begin{tabular}{ccccccc}
\toprule
$\lambda$ in training        & $\lambda$ in testing & \begin{tabular}[c]{@{}c@{}}3D mAP\\ (fog level = 0)\end{tabular} & \begin{tabular}[c]{@{}c@{}}3D mAP\\ (fog level = 1)\end{tabular} & \begin{tabular}[c]{@{}c@{}}3D mAP\\ (fog level = 2)\end{tabular} & \begin{tabular}[c]{@{}c@{}}3D mAP\\ (fog level = 3)\end{tabular} & \begin{tabular}[c]{@{}c@{}}3D mAP\\ (fog level = 4)\end{tabular} \\ \midrule
\multirow{4}{*}{0.1} & 0.1          & 77.56                                                            & 77.03                                                            & 62.40                                                            & 50.11                                                            & 25.27                                                            \\
                     & 0.2          & 75.92                                                            & 75.79                                                            & 61.09                                                            & 49.00                                                            & 25.28                                                            \\
                     & 0.3          & 75.46                                                            & 74.46                                                            & 60.84                                                            & 48.84                                                            & 25.72                                                            \\
                     & 0.5          & 73.57                                                            & 71.79                                                            & 59.10                                                            & 47.15                                                            & 24.99                                                            \\ \midrule\midrule
\multirow{4}{*}{0.2} & 0.1          & 77.59                                                            & 77.43                                                            & 64.06                                                            & 52.57                                                            & 23.90                                                            \\
                     & 0.2          & 77.21                                                            & 76.63                                                            & 62.94                                                            & 51.10                                                            & 23.99                                                            \\
                     & 0.3          & 75.84                                                            & 75.23                                                            & 62.15                                                            & 50.44                                                            & 23.97                                                            \\
                     & 0.5          & 73.73                                                            & 72.61                                                            & 59.41                                                            & 48.87                                                            & 22.40                                                            \\ \midrule\midrule
\multirow{4}{*}{0.3} & 0.1          & 79.51                                                            & 78.77                                                            & 63.78                                                            & 52.95                                                            & 25.94                                                            \\
                     & 0.2          & \textbf{79.80}                                                   & \textbf{78.84}                                                   & \textbf{64.73}                                                   & 53.26                                                            & \textbf{28.87}                                                   \\
                     & 0.3          & 79.67                                                            & 77.91                                                            & 63.33                                                            & 52.02                                                            & 26.28                                                            \\
                     & 0.5          & 76.71                                                            & 75.75                                                            & 61.28                                                            & 51.56                                                            & 26.46                                                            \\ \midrule\midrule
\multirow{4}{*}{0.5} & 0.1          & 77.18                                                            & 76.67                                                            & 62.35                                                            & 51.45                                                            & 24.30                                                            \\
                     & 0.2          & 78.91                                                            & 77.87                                                            & 63.61                                                            & \textbf{53.49}                                                   & 28.49                                                            \\
                     & 0.3          & 79.47                                                            & 78.35                                                            & 63.57                                                            & 52.22                                                            & 27.19                                                            \\
                     & 0.5          & 78.57                                                            & 77.12                                                            & 62.47                                                            & 51.18                                                            & 26.29                                                            \\ \bottomrule
\end{tabular}
\caption{Performance with different hyperparameters $\lambda$ in FAD both in the training and testing stages. }
\label{lambda}
\end{table*}

\subsection{Discussion about the Counter-Intuitive Results on K-Radar dataset}
As shown by the results in the main paper: K-Radar inverse weather dataset experiment, all methods perform better in many severe weather conditions (e.g., cloudy, foggy, etc.) than in normal weather. After analyzing the data and experimental results, a possible main reason is the distribution differences of labeled objects in different weather conditions.We will verify this in two respects.

\paragraph{The performance of 4D Radar-based method in different weather condition.} Benefiting from the sensor weather robustness of 4D radar, the 4D Radar-based method receives little performance impact from weather. Therefore, the performance of the 4D Radar-based method under each weather can be used to determine the difference in difficulty caused by the distribution of labeled objects for different weather on the K-radar dataset. As shown in the results demonstrated by the 4D Radar-based method (RTNH, PointPillars) in Tables \ref{tab:full_kradar_v1.0} and \ref{v2.1}, it can be found that the performance of the 4D Radar-based method is significantly worse than that of most of the bad weather in normal weather. This result illustrates that labeled objects in bad weather scenarios are much easier to detect in the K-Radar dataset. This makes it possible that even when LiDAR is degraded in bad weather, the performance is still higher than in normal weather due to the fact that the labeled targets in the scene are easier to detect.

\begin{figure}[!h]
    
    \includegraphics[width=\columnwidth]{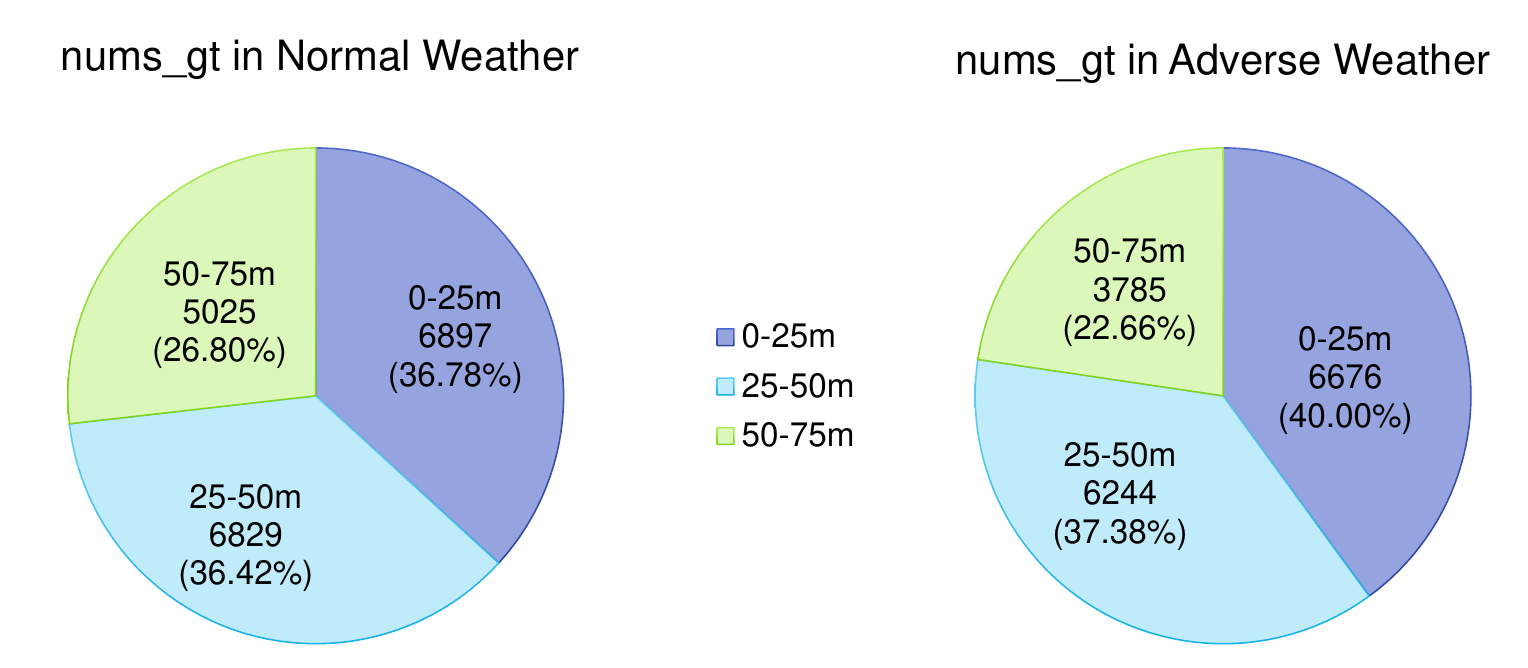}
  \centering
  \caption{Distribution of labeled bounding boxes in normal (left) and severe (right) weather in the K-Radar dataset.}
  \label{num_gt}
\end{figure}

\paragraph{Labeled Object Distance-Number Distribution of K-Radar Dataset.} We counted the number of labeled objects  at different distances by weather category (normal, severe), as shown in Figure \ref{num_gt}. It can be found that under normal weather, the proportion of the number of labeled objects (26.80\%)  at long distances (50-75m) is greater than that under severe weather (22.66\%), while the proportion of the number of labeled objects  (36.78\%) at closed distances (0-25m) is smaller than that under severe weather (40.0\%). This data reflects the difficulty of the scenarios in different weather, explaining the counter-intuitive results on K-Radar dataset.

\subsection{Experiments of the Hyperparameter $\lambda$ in FAD}
We have conducted sufficient experimental discussion on the hyperparameter $\lambda$ in FAD both in the training and testing stages. The experimental results are shown in Table \ref{lambda}, which shows that too small $\lambda$ will lead to too much noise residue and an insignificant denoising effect, while too large $\lambda$ will lose a large number of foreground points affecting the detection of the object. Moreover, the performance degree of different $\lambda$ under different fog levels is also different, which is due to the different importance of 4D radar under different fog levels. In the end, we chose the setting with the best overall performance with $\lambda$ = 0.3 for training and $\lambda$ = 0.2 for testing, which is also in line with our expectations. Firstly, $\lambda$ cannot be used with the 0.5 threshold for conventional binary classification, which needs to be appropriately lowered. Secondly, the threshold $\lambda$ for training should be slightly higher than that for testing due to the increased number of foreground points caused by the data augmentations, such as Ground Truth Sampling.

\begin{table*}[!h]
    \centering
    \begin{tabular}{cccc|cccccccc}
    \hline
        MME & FAD & \{IM\}\^2 & MSGF & Total & Normal & Overcast & Fog & Rain & Sleet & Light Snow & Heavy Snow \\ \hline
        ~ & ~ & ~ & ~ & 37.8 & 39.8 & 46.3 & 59.8 & 28.2 & 31.4 & 50.7 & 24.6 \\ 
        \checkmark & ~ & ~ & ~ & 51.6 & 46.1 & 58.7 & 77.4 & 52.7 & 43.6 & 63.5 & 42.8 \\ 
        \checkmark & \checkmark & ~ & ~ & 51.8 & 49.8 & 60.8 & 77.2 & 50.9 & 43.7 & 64.5 & 33.0 \\ 
        \checkmark & \checkmark & \checkmark & ~ & 52.5 & 51.8 & 63.6 & 66.7 & 51.6 & 46.0 & 52.1 & 36.6 \\ 
        \checkmark & \checkmark & \checkmark & \checkmark & 53.5 & 53.0 & 64.1 & 73.2 & 53.8 & 46.2 & 52.4 & 37.0 \\ \hline
    \end{tabular}
    
    \caption{The experimental ablation results of each module on K-Radar dataset. Best in \textbf{bold}.}
    \label{ablaK}
\end{table*}

\subsection{More Implement Details}
For the training strategy, we train the entire network with the loss of 30 epochs. We use Adam optimizer with lr= 1e-3, $\beta$1 = 0.9, $\beta$2 = 0.999. 

For the K-Radar dataset, we preprocess the 4D radar sparse tensor by selecting only the top 10240 points with high power measurement. We present the set the point cloud range as [0m, 72m] for the X axis, [6.4m, 6.4m] for the Y axis, and [-2m, 6m] for the Z axis setting the same environment with version 1.0 K-Radar. And [0m, 72m] for the X axis, [-16m, 16m] for the Y axis, and [-2m, 7.6m] for the Z axis setting the same environment with version 2.1 K-Radar. The voxel size is set to (0.4m, 0.4m, 0.4m). 

For the VoD dataset, following KITTI \citep{kitti}, we calculate the 3D Average Precision (3D AP) across 40 recall thresholds (R40) for different classes.  Also, following VoD's \citep{vod} evaluation metrics, we calculate class-wise AP and mAP averaged over classes. The calculation encompasses the entire annotated region (camera FoV up to 50 meters) and the "Driving Corridor" region ([-4 m < x < +4 m, z < 25 m]). For both KITTI metrics and VoD metrics, for AP calculations, we used an IoU threshold specified in VoD, requiring a 50\% overlap for car class and 25\% overlap for pedestrian and cyclist classes. 

\subsection{More Performance on K-Radar Dataset}

We also have conducted more ablation on K-Radar dataset as in Table \ref{ablaK}. The first row represents the RTNH as the LiDAR-only baseline for comparison.
Adding MME multi-modal information significantly improves performance, especially in challenging weather. Including all modules further enhances total performance. However, {IM}² causes a notable performance drop in fog and snow, likely due to feature redundancy. While MSGF mitigates this to some extent, it remains a concern and will be a key area for future optimization. These results and analyses will be included in the supplementary materials.

The main text is bound by space constraints and only results using IoU=0.5 and v1.0 labeling are shown on K-Radar. Here we additionally show results using IoU=0.3 with v1.0 labels as in Table. \ref{tab:full_kradar_v1.0} and results using IoU=0.3 with v2.0 labels as in Table. \ref{v2.1}. The experimental results all demonstrate the superior performance of our L4DR.

\begin{table*}[!h]

\centering
    \setlength{\tabcolsep}{3.pt}
\small
\begin{tabular}{cc|cccccccccc}
\hline
{{Methods}}                                           & {{Modality}}         & {{IoU}} & {{Metric}} & {{Total}} & {{Normal}} & {{Overcast}} & {{Fog}} & {{Rain}} & {{Sleet}} & {{Lightsnow}} & {{Heavysnow}} \\ \hline\hline
\multirow{4}{*}{\shortstack{RTNH\\(NeurIPS 2022)}}   & \multirow{4}{*}{4DR}
& \multirow{2}{*}{0.5}  & $AP_{BEV}$ &36.0 &35.8 &41.9 &44.8 &30.2 &34.5 &63.9 &55.1 \\ 
& &   & $AP_{3D}$ &14.1 &19.7 &20.5 &15.9 &13.0 &13.5 &21.0 &6.36  \\ \cline{3-12}
&   & \multirow{2}{*}{0.3} & $AP_{BEV}$ & 41.1  & 41.0  & 44.6 & 45.4 & 32.9 & 50.6 & 81.5  & 56.3  \\
& &  & $AP_{3D}$ &37.4 &37.6 &42.0 &41.2 &29.2 &49.1 &63.9 &43.1   \\ \hline
\multirow{4}{*}{\shortstack{PointPillars\\(CVPR 2019)}}   & \multirow{4}{*}{L}
& \multirow{2}{*}{0.5}  & $AP_{BEV}$ &49.1 &48.2 &53.0 &45.4 &44.2 &45.9 &74.5 &53.8 \\
& &   & $AP_{3D}$ &22.4 &21.8 &28.0 &28.2 &27.2 &22.6 &23.2 &12.9  \\ \cline{3-12}
&   & \multirow{2}{*}{0.3} & $AP_{BEV}$ &51.9 &51.6 &53.5 &45.4 &44.7 &54.3 &81.2 &55.2  \\ 
& &   & $AP_{3D}$ &47.3 &46.7 &51.9 &44.8 &42.4 &45.5 &59.2 &\underline{55.2}  \\ \hline
\multirow{4}{*}{\shortstack{RTNH\\(NeurIPS 2022)}}   & \multirow{4}{*}{L}
& \multirow{2}{*}{0.5}  & $AP_{BEV}$ &66.3 &65.4 &87.4 &83.8 &73.7 &\underline{48.8} &78.5 &48.1 \\
& &   & $AP_{3D}$ &37.8 &39.8 &46.3 &\underline{59.8} &28.2 &\underline{31.4} &50.7 &24.6  \\ \cline{3-12}
&   & \multirow{2}{*}{0.3} & $AP_{BEV}$ &76.5 &76.5 &88.2 &86.3 &77.3 &55.3 &81.1 &59.5 \\ 
& &  & $AP_{3D}$  &72.7 &73.1 &76.5 &84.8 &64.5 &\underline{53.4} &80.3 &52.9 \\ \hline
\multirow{4}{*}{\shortstack{InterFusion\\(IROS 2023)}}   & \multirow{4}{*}{L+4DR}
& \multirow{2}{*}{0.5}  & $AP_{BEV}$ &52.9 &50.0 &59.0 &80.3 &50.0 &22.7 &72.2 &53.3 \\
& &   & $AP_{3D}$ &17.5 &15.3 &20.5 &47.6 &12.9 &9.33 &\underline{56.8} &25.7  \\ \cline{3-12}
&   & \multirow{2}{*}{0.3} &$AP_{BEV}$ &57.5 &57.2 &60.8 &81.2 &52.8 &27.5 &72.6 &57.2 \\ 
& &   & $AP_{3D}$ &53.0 &51.1 &58.1 &80.9 &40.4 &23.0 &71.0 &55.2  \\ \hline
\multirow{4}{*}{\shortstack{3D-LRF\\(CVPR 2024)}}   & \multirow{4}{*}{L+4DR}
& \multirow{2}{*}{0.5}  & $AP_{BEV}$ &\underline{73.6} &\underline{72.3} &\underline{88.4} &\underline{86.6} &\underline{76.6} &47.5 &\underline{79.6} &\textbf{64.1} \\
& &   & $AP_{3D}$ &\underline{45.2} &\underline{45.3} &\underline{55.8} &51.8 &\underline{38.3} &23.4 &\textbf{60.2} &\underline{36.9}  \\ \cline{3-12}
&   &\multirow{2}{*}{0.3} & $AP_{BEV}$ &\textbf{84.0} &\underline{83.7} &\underline{89.2} &\textbf{95.4} &\underline{78.3} &\underline{60.7} &\underline{88.9} &\textbf{74.9} \\ 
& &  & $AP_{3D}$ &74.8 &\textbf{81.2} &\textbf{87.2} &\underline{86.1} &\underline{73.8} &49.5 &\textbf{87.9} &\textbf{67.2}  \\ \hline
\multirow{4}{*}{\shortstack{L4DR\\(Ours)}}   & \multirow{4}{*}{L+4DR}
& \multirow{2}{*}{0.5}  & $AP_{BEV}$ &\textbf{77.5} &\textbf{76.8} &\textbf{88.6} &\textbf{89.7} &\textbf{78.2} &\textbf{59.3} &\textbf{80.9} &\underline{53.8} \\
& &                     & $AP_{3D}$  &\textbf{53.5} &\textbf{53.0} &\textbf{64.1} &\textbf{73.2} &\textbf{53.8} &\textbf{46.2} &52.4 &\textbf{37.0}  \\ \cline{3-12}
& &\multirow{2}{*}{0.3} & $AP_{BEV}$ &\underline{79.5} &\textbf{86.0} &\textbf{89.6} &\underline{89.9} &\textbf{81.1} &\textbf{62.3} &\textbf{89.1} &\underline{61.3} \\ 
& &                     & $AP_{3D}$  &\textbf{78.0} &\underline{77.7} &\underline{80.0} &\textbf{88.6} &\textbf{79.2} &\textbf{60.1} &\underline{78.9} &51.9  \\ \hline
\end{tabular}
\caption{Quantitative results of different 3D object detection methods on K-Radar dataset. We present the modality of each method (L: LiDAR, 4DR: 4D radar) and detailed performance for each weather condition. Best in \textbf{bold}, second in \underline{underline}. }
\label{tab:full_kradar_v1.0}

 \end{table*}

\begin{table*}[!h]
\centering
\small
\begin{tabular}{ccccccccccc}
\toprule
\multirow{2}{*}{\textbf{Class}}                                          & \multirow{2}{*}{\textbf{Method}}         & \multirow{2}{*}{\textbf{Modality}} & \multirow{2}{*}{\textbf{Total}} & \multirow{2}{*}{\textbf{Normal}} & \multirow{2}{*}{\textbf{Li. Snow}} & \multirow{2}{*}{\textbf{He. Snow}} & \multirow{2}{*}{\textbf{Rain}}         & \multirow{2}{*}{\textbf{Sleet}}                 & \multirow{2}{*}{\textbf{Overcast}}     & \multirow{2}{*}{\textbf{Fog}}                   \\ 
\\ \midrule
\multirow{6}{*}{Sedan}                                                    & Pointpillars* (CVPR2019)                 & 4DR                  &42.8 &35.0              & \multicolumn{1}{c}{53.6}                                                        & \multicolumn{1}{c}{48.3}                                                        & \multicolumn{1}{c}{37.4}                  & \multicolumn{1}{c}{37.5}                           & \multicolumn{1}{c}{53.9}                  & 77.3                           \\
  & RTNH(NIPS2022)                           & 4DR          &48.2 &35.5                      & \multicolumn{1}{c}{65.6}                                                        & \multicolumn{1}{c}{52.6}                                                        & \multicolumn{1}{c}{40.3}                  & \multicolumn{1}{c}{48.1}                           & \multicolumn{1}{c}{58.8}                  & 79.3                           \\  
  & Pointpillars* (CVPR2019)                 & L  &69.7 &68.1  & \multicolumn{1}{c}{79.0}                                                        & \multicolumn{1}{c}{51.5}                                                        & \multicolumn{1}{c}{77.7}                  & \multicolumn{1}{c}{59.1}                           & \multicolumn{1}{c}{79.0}                  & 89.2                           \\  
  & \multirow{1}{*}{InterFusion* (IROS2022)} & \multirow{1}{*}{L+4DR}     &69.9 &69.0   & \multicolumn{1}{c}{\multirow{1}{*}{79.1}}                                       & \multicolumn{1}{c}{\multirow{1}{*}{51.7}}                                       & \multicolumn{1}{c}{\multirow{1}{*}{77.1}} & \multicolumn{1}{c}{\multirow{1}{*}{58.9}}          & \multicolumn{1}{c}{\multirow{1}{*}{77.9}} & \multirow{1}{*}{\textbf{89.5}} \\\cmidrule(l){2-11}
  & L4DR (Ours)                                     & L+4DR      &\textbf{75.8} &\textbf{74.6}                        & \multicolumn{1}{c}{\textbf{87.5}}                                               & \multicolumn{1}{c}{\textbf{58.4}}                                               & \multicolumn{1}{c}{\textbf{77.8}}         & \multicolumn{1}{c}{\textbf{61.4}}                  & \multicolumn{1}{c}{\textbf{79.2}}         & 89.3                           \\ \midrule \midrule
\multirow{6}{*}{\begin{tabular}[c]{@{}c@{}}Bus\\ or\\ Truck\end{tabular}} & Pointpillars* (CVPR2019)                 & 4DR &29.4 &25.8   & \multicolumn{1}{c}{64.1}                                                        & \multicolumn{1}{c}{34.9}                                                        & \multicolumn{1}{c}{0.0}                   & \multicolumn{1}{c}{18.0}                           & \multicolumn{1}{c}{21.5}                  & -                              \\
  & RTNH(NIPS2022)                           & 4DR         &34.4&25.3                       & \multicolumn{1}{c}{78.2}                                                        & \multicolumn{1}{c}{46.3}                                                        & \multicolumn{1}{c}{0.0}                     & \multicolumn{1}{c}{28.5}                           & \multicolumn{1}{c}{31.1}                  & -                              \\  
  & Pointpillars* (CVPR2019)                 & L      &53.8  &52.9                           & \multicolumn{1}{c}{84.1}                                                        & \multicolumn{1}{c}{50.7}                                                        & \multicolumn{1}{c}{3.7}                   & \multicolumn{1}{c}{61.8}                           & \multicolumn{1}{c}{77.3}                  & -                              \\  
  & \multirow{1}{*}{InterFusion* (IROS2022)} & \multirow{1}{*}{L+4DR}     &56.9 &56.2               & \multicolumn{1}{c}{\multirow{1}{*}{\textbf{85.7}}}                              & \multicolumn{1}{c}{\multirow{1}{*}{40.5}}                                       & \multicolumn{1}{c}{\multirow{1}{*}{6.4}}  & \multicolumn{1}{c}{\multirow{1}{*}{\textbf{70.6}}} & \multicolumn{1}{c}{\multirow{1}{*}{80.5}} & \multirow{1}{*}{-}             \\\cmidrule(l){2-11}
  & L4DR (Ours)                                     & L+4DR          &59.7 &59.4                             & \multicolumn{1}{c}{84.4}                                                        & \multicolumn{1}{c}{\textbf{51.9}}                                               & \multicolumn{1}{c}{\textbf{8.1}}          & \multicolumn{1}{c}{66.1}                           & \multicolumn{1}{c}{\textbf{86.4}}         & -                              \\ \hline

\end{tabular}
\caption{Performance Comparing L4DR under each type of real-world weather condition. The best performances are highlighted in \textbf{bold}. * indicates our reproduction using open-source code from original authors. - indicates that there is no object or the original author does not report performance. L indicates LiDAR and 4DR indicates 4D radar. }
\label{v2.1}

\end{table*}

\begin{figure*}[!h]
    
    \includegraphics[width=0.8\textwidth]{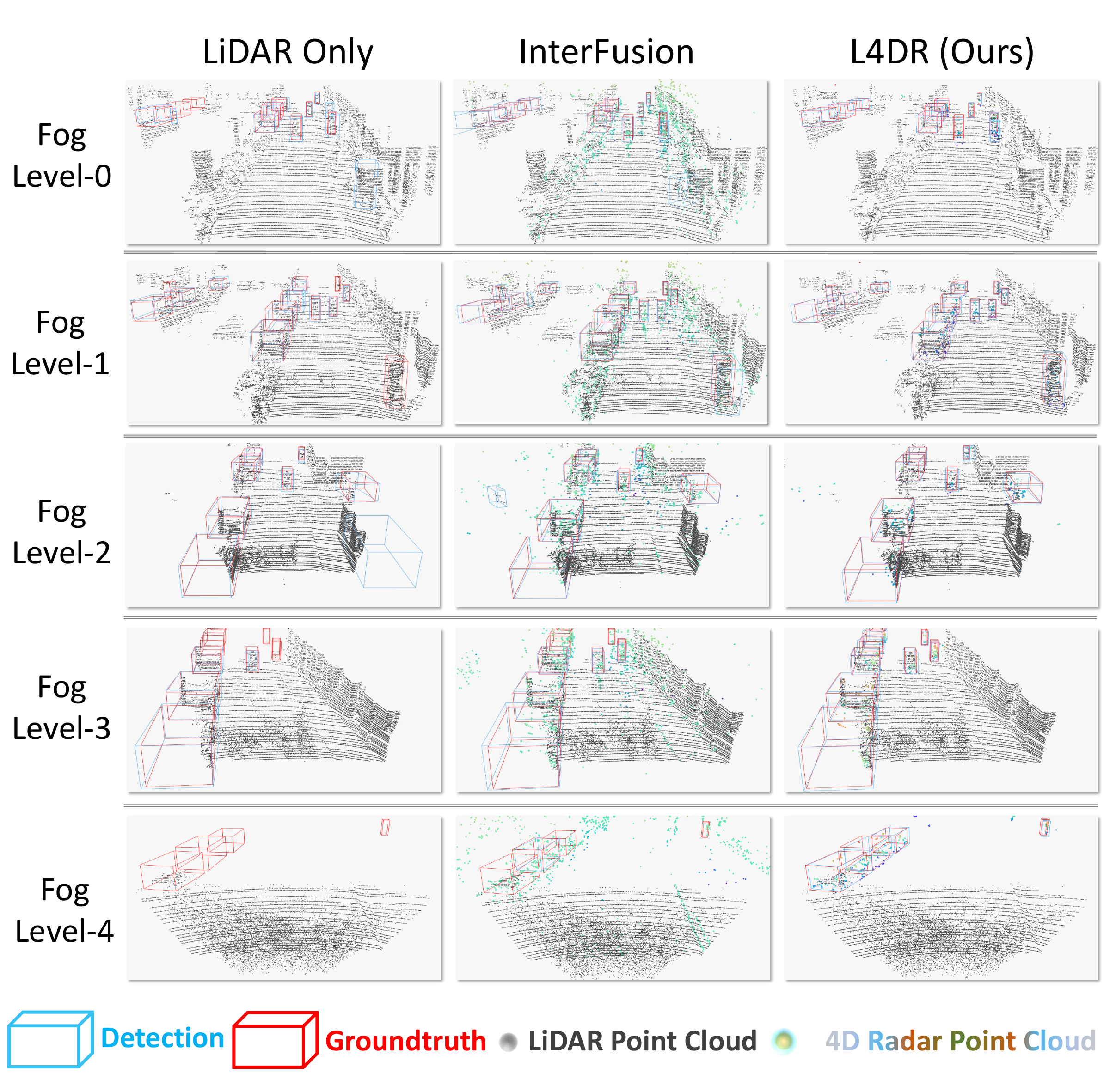}
  \centering
  \caption{Visualization performance comparison. We visualize object detection results with LiDAR-only (left), InterFusion (middle), and our L4DR (right) on the VoD-fog dataset with different simulated fog levels (0-4, from up to bottom). The red and blue 3D bounding boxes indicate groundtruths and model predictions, respectively, the grey points are LiDAR point clouds and the colored points are 4D radar point clouds. Our L4DR shows the 4D radar point cloud after denoising by the FAD module.}
  \label{vis_sup}
\end{figure*}

\begin{table*}[!h]
    \centering
    
    \begin{tabular}{l|l|l|l|l|l|l}
    \hline
        MME & FAD & \{IM\}\^2 & MSGF & GFLOPs & Inference Speed & Params \\ \hline
        - & - & - & - & 88.19 & 25ms & 18.04M \\ \hline
        \checkmark  & ~ & ~ & ~ & 88.23 (+0.04) & 71ms (+46ms) & 18.04M (+0M) \\ \hline
        \checkmark  & \checkmark  & ~ & ~ & 91.89 (+3.66) & 80ms (+9ms) & 20.41M (+2.37M) \\ \hline
        \checkmark  & \checkmark  & \checkmark  & ~ & 232.95 (+141.06) & 91ms  (+9ms) & 55.13M (+34.72M) \\ \hline
        \checkmark  & \checkmark  & \checkmark  & \checkmark  & 255.60 (+22.65) & 96ms (+5ms) & 61.32M (+6.19M) \\ \hline
    \end{tabular}
    \caption{The ablation experiments on model computation FLOPs, inference speed, and model parameters for each module.}
    \label{over}
\end{table*}

\subsection{Computational Time and Overhead}
We conducted ablation experiments on model computation FLOPs, inference speed, and model parameters for each module. As shown in the table \ref{over}, the first row represents the baseline overhead for LiDAR-4DRadar fusion using the naive concatenation of LiDAR and 4DRadar BEV features. Notably, the MME module significantly increases inference time due to computations between the two modal voxels, while the \{IM\}$^2$ module substantially increases GFLOPs and parameters due to the addition of an extra backbone branch.

\subsection{Experimental Visualization Results}

To better visualize how our method improves detection performance, we compare our L4DR with InterFusion \citep{InterFusion} under different simulated fog levels, as shown in Figure \ref{vis_sup}. Our L4DR effectively filters out a substantial amount of noise in 4D radar points (depicted as colored points). Furthermore, our L4DR achieves an effective fusion of LiDAR and 4D radar to increase the precise recall of hard-to-detect objects and reduce false detections.

\bibliography{aaai25}

\end{document}